\documentclass{article}

\usepackage{arxiv}

\usepackage[utf8]{inputenc} 
\usepackage[T1]{fontenc}    
\usepackage{hyperref}       
\usepackage{url}            
\usepackage{booktabs}       
\usepackage{amsfonts}       
\usepackage{nicefrac}       
\usepackage{microtype}      
\usepackage{lipsum}
\usepackage{graphicx}
\usepackage{tikz}
\usepackage{amsmath}
\usepackage{multirow}
\usetikzlibrary{arrows.meta,positioning,shapes.geometric}
\graphicspath{ {./images/} }

\title{A Discordance-Aware Multimodal Framework with Multi-Agent Clinical Reasoning}

\author{
 Pegah Ahadian \\
  Department of Computer Science\\
  Kent State University\\
  Ohio, USA \\
  \texttt{pahadian@kent.edu} \\
   \And
    Mingrui Yang \\
  Department of Biomedical Engineering\\
  Cleveland Clinic\\
  Ohio, USA \\
  \texttt{yangm@ccf.org} \\
  \And
 Sixu Chen \\
  Department of Computer Science\\
  Kent State University\\
  Ohio, USA \\
  \texttt{schen53@kent.edu} \\
     \And
    Xiaojuan Li \\
  Department of Biomedical Engineering\\
  Cleveland Clinic\\
  Ohio, USA \\
  \texttt{lix6@ccf.org} \\
  \And
 Qiang Guan \\
  Department of Computer Science\\
  Kent State University\\
  Ohio, USA \\
  \texttt{qguan@kent.edu} \\
}

\begin{document}
\maketitle
\begin{abstract}

\textbf{Background:}
Knee osteoarthritis (OA) frequently exhibits discordance between structural damage observed in imaging and patient-reported symptoms such as pain. This mismatch complicates clinical interpretation and patient stratification and remains insufficiently modeled in existing decision-support systems.

\textbf{Methods:}
We propose a discordance-aware multimodal framework that combines machine learning prediction models with a tool-grounded multi-agent reasoning system. Using baseline data from the FNIH Osteoarthritis Biomarkers Consortium (600 knees), we trained multimodal models to predict two progression tasks: (i) joint-space-loss-only progression versus non-progression and (ii) pain-only progression versus non-progression. The predictive system integrates three modality-specific experts: a CatBoost tabular model using demographic, radiographic, MRI-derived scalar, and biomarker features; MRI image embeddings extracted using a ResNet18 backbone; and X-ray embeddings derived from the same architecture. Expert predictions are fused using a stacking ensemble. Residual-based models estimate expected pain from structural features, enabling the computation of a pain–structure discordance score between observed and expected symptoms. A multi-agent reasoning layer interprets these signals to assign clinically interpretable OA phenotypes and generate phenotype-specific management recommendations.

\textbf{Results:}
Using 5-fold stratified cross-validation with out-of-fold evaluation, the full multimodal stacking model combining tabular variables, MRI and X-ray embeddings, and biochemical biomarkers achieved the best performance. For the JSL-only progression task, the model achieved AUC 0.702. For pain-only progression, the model achieved AUC 0.611. Imaging embeddings alone provided limited predictive signal, whereas clinically interpretable radiographic and MRI scalar features contributed stronger discrimination. Multimodal fusion improved performance by integrating complementary structural and biochemical information.

\textbf{Conclusions:}
Multimodal fusion of tabular clinical variables, imaging-derived features, deep image embeddings, and biochemical biomarkers improves structural progression prediction in knee OA. By coupling this prediction layer with explicit pain–structure discordance modeling and a tool-grounded multi-agent reasoning framework, the proposed architecture supports interpretable phenotype assignment and structured clinical decision support for osteoarthritis management.
\end{abstract}

\section{Introduction}

Knee osteoarthritis (OA) is a heterogeneous chronic musculoskeletal disease in which structural degeneration, pain, and functional impairment often evolve along partially independent trajectories \cite{b13, b18, b19}. In routine care, clinicians are frequently confronted with patients who report substantial pain despite limited radiographic damage, as well as patients with marked structural progression but comparatively mild symptoms. This discordance complicates diagnosis, prognostic stratification, and treatment planning because conventional clinical reasoning often assumes a tighter alignment between imaging severity and symptom burden than is observed in practice \cite{b1, b5, b7}.

Recent advances in multimodal machine learning have created new opportunities to integrate heterogeneous OA data sources \cite{b9, b10, b11, b16}, including demographic covariates, radiographic features, MRI-derived quantitative measures, biochemical biomarkers \cite{b8}, and deep image embeddings. At the same time, most existing OA prediction pipelines remain optimized for a single endpoint and typically collapse structural and symptomatic information into one predictive target. Such formulations may improve benchmark metrics, but they obscure an important clinical question: whether the patient's symptoms are concordant with, or disproportionate to, the level of structural disease. In OA, that distinction is clinically meaningful because it may indicate pain sensitization, early structural disease, compensatory adaptation, or other non-structural drivers of disability.

In parallel, multi-agent large language model systems have emerged as a promising design pattern for complex clinical reasoning \cite{b4}. AutoGen formalizes this idea as a framework of conversable and customizable agents with unified send/receive interfaces, auto-reply mechanisms, and flexible conversation programming, including both static and dynamic multi-agent interaction patterns \cite{b2}. These properties make multi-agent systems appealing for tasks in which different evidence streams must be interpreted separately and then reconciled through explicit dialogue rather than collapsed into a single monolithic response \cite{b12, b14, b15, b16}.

Within knee osteoarthritis specifically, KOM demonstrated that a multi-agent architecture can support broader KOA management workflows through an Assessment Agent, a Risk Agent, and a Therapy Agents Group, and that clinician-AI collaboration can improve workflow efficiency and plan quality in simulated clinical settings \cite{b3}. However, KOM is designed for end-to-end KOA evaluation and treatment planning. It does not make pain-structure discordance the primary representation to be detected, debated, and interpreted.

This study addresses that gap. We propose a discordance-aware multimodal framework in which structural progression prediction and symptom expectation modeling are explicitly separated. First, we develop multimodal predictive models for divided progression tasks using tabular features, radiographic and MRI-derived scalar measurements, deep MRI and X-ray embeddings, and biochemical biomarkers. Second, we estimate expected symptom burden from structural information and define discordance scores as residuals between observed and structure-expected pain or function. Third, we introduce a tool-grounded multi-agent reasoning layer in which specialized agents discuss whether structural risk and symptom burden are concordant or discordant, and then assign phenotype-level interpretations and management recommendations.

The central hypothesis is that OA decision support should not merely predict progression, but should explicitly reason about disagreement between structural and symptomatic signals. Under this view, discordance is not residual noise. It is a clinically informative signal that can drive phenotype assignment, management prioritization, and auditably structured discussion among specialist agents.

Our contributions are fourfold. First, we present a multimodal progression prediction framework for divided OA progression tasks using a mixture-of-experts design. Second, we formalize pain-structure discordance using residual-based symptom expectation models. Third, we design a multi-agent conversation protocol in which structuralist, physiologist, and lead consultant agents synthesize concordant and discordant evidence. Fourth, we provide a clinician-grounded evaluation setup comparing deterministic, single-agent, and decomposed multi-agent reporting configurations, with specialist-output specifications and blinded clinician review packets prepared for supplementary analysis.

\section{Methods}

\subsection{Data sources and cohort definition}

The source Clinical\_FNIH table contains 600 knees \cite{b8}. After exclusion of knees with missing predictors required for the progression modeling pipeline, 545 knees remained in the structural modeling cohort. A smaller complete-case subset of 90 knees with all variables required for residual-based discordance estimation was used for symptom expectation modeling and threshold analysis, table \ref{tab:fnih_clinical_char}.

Each record corresponds to a knee-level observation. Baseline predictors include demographic variables, radiographic structural indicators, MRI-derived scalar features, and biochemical markers. In addition, MRI and X-ray images were processed to derive deep image embeddings used in multimodal fusion experiments.

\begin{table}[ht]
\centering
\small
\caption{Baseline demographic and clinical characteristics of the FNIH OA Biomarkers (Clinical\_FNIH dataset \cite{b8}).}
\label{tab:fnih_clinical_char}
\begin{tabular}{lcc}
\toprule
Characteristic & Category / value & Frequency (\%) \\
\midrule
Knees (records) &
Total knees in Clinical\_FNIH &
600 (100.0)  \\[0.4ex]

Side &
Right knee &
322 (53.7)  \\
&
Left knee &
278 (46.3) \\[0.8ex]

CASE / GROUPTYPE &
JSL and pain progressor (Case = 1) &
194 (32.3) \\
&
JSL only progressor (Case = 2) &
103 (17.2)  \\
&
Pain only progressor (Case = 3) &
103 (17.2)  \\
&
Non–progressor (Case = 4) &
200 (33.3)  \\[0.8ex]

Baseline Kellgren–Lawrence grade
(V00XRKL; 0–4) &
Grade 1 &
75 (12.5)  \\
&
Grade 2 &
306 (51.0) \\
&
Grade 3 &
219 (36.5) \\[0.8ex]

Baseline medial JSN
(V00XRJSM; OARSI 0–3) &
Grade 0 &
162 (27.0)  \\
&
Grade 1 &
219 (36.5)  \\
&
Grade 2 &
219 (36.5) \\[0.8ex]

Baseline lateral JSN
(V00XRJSL; OARSI 0–3) &
Grade 0 &
587 (97.8)  \\
&
Grade 1 &
13  (2.2)  \\[0.8ex]

Race by sex\footnotemark &
White or Caucasian, male &
210 (35.0 of 600)  \\
&
White or Caucasian, female &
265 (44.2 of 600) \\
&
Non–white, male &
37 (6.2 of 600)  \\
&
Non–white, female &
88 (14.7 of 600) \\[0.8ex]

Age (years) by sex &
45–49, male / female &
25 / 28  \\
&
50–59, male / female &
92 / 120  \\
&
60–69, male / female &
66 / 132  \\
&
70–79, male / female &
64 / 73  \\
\bottomrule
\end{tabular}
\end{table}

\footnotetext{Race counts are reported exactly as in the Clinical\_FNIH descriptive documentation; percentages in this table are calculated relative to 600 knees.}

\subsection{Prediction tasks}

To explicitly model the separation between structural deterioration and symptom worsening in knee osteoarthritis, we formulated two divided binary progression tasks rather than collapsing all progression patterns into a single endpoint. This design choice reflects the central clinical premise of the study: structural progression and pain progression are related but non-equivalent processes, and a clinically useful decision-support framework should preserve that distinction.

The first task, \emph{JSL Only vs Non}, classifies knees exhibiting isolated joint-space-loss progression against non-progressor knees. In this setting, the positive class corresponds to knees with structural worsening in the absence of concurrent pain progression, whereas the negative class comprises knees without progression. This task is intended to capture a predominantly structure-driven trajectory and provides a direct test of whether baseline radiographic, MRI-derived, and biomarker features can predict future structural decline even when symptom escalation is not prominent. From the perspective of the proposed framework, this task is especially important because it operationalizes a phenotype in which tissue-level degeneration may advance despite relatively limited symptomatic change.

The second task, \emph{Pain Only vs Non}, classifies knees with isolated pain progression against non-progressor knees. Here, the positive class represents knees whose symptom burden worsens over time without parallel structural progression, while the negative class again consists of knees without progression. This task is clinically complementary to the JSL-only setting because it isolates a symptom-dominant trajectory. Such cases are particularly relevant for discordance-aware modeling, as they may reflect pain sensitization, inflammatory fluctuations, psychosocial contributors, altered pain processing, or disease mechanisms not adequately captured by conventional structural imaging markers.

These divided tasks were selected instead of a single aggregated progression label because a unified endpoint would obscure the very mismatch that motivates the present work. If structural progression and pain progression are merged into one outcome, the model may achieve acceptable discrimination while remaining uninformative about whether the predicted risk is driven by structural change, symptom escalation, or both. In contrast, the separated formulation allows us to examine whether multimodal inputs carry different predictive signal for structural and symptomatic trajectories. This distinction is necessary for the downstream discordance layer, where the system must assess whether a patient's observed pain profile is concordant with, or disproportionate to, the level of structural risk.

From a modeling perspective, the divided-task setup also provides a cleaner basis for interpreting feature contributions. Structural descriptors such as joint-space narrowing, Kellgren--Lawrence grade, MRI-derived morphometric measures, and biochemical markers are expected to be more informative for the JSL-only task than for pain-only progression. Conversely, weaker performance on the pain-only task would support the hypothesis that symptom progression is not fully determined by structural burden. This asymmetry is not a weakness of the framework. Rather, it is part of the scientific question being tested, namely whether discordance between structural and symptomatic progression can be made explicit and clinically interpretable.

Finally, defining separate progression tasks creates a principled bridge to the multi-agent reasoning layer. The structural prediction branch provides evidence about future tissue-level worsening, while the symptom-related branch informs whether pain escalation follows a distinct trajectory. The downstream agents can then reason over these partially independent signals rather than relying on a single conflated risk score. In this way, the prediction task design is not merely a technical preprocessing decision, but a foundational part of the proposed discordance-aware framework.

\subsection{Multimodal progression prediction}

To evaluate the predictive value of clinical variables, imaging-derived features, and biochemical biomarkers, we constructed a multimodal mixture-of-experts model with three modality-specific predictors.

\paragraph{Tabular expert (T).}
The tabular expert is a CatBoost model trained on demographic variables, radiographic scalar features, MRI-derived scalar measurements, and, in the full configuration, biochemical biomarkers measured from serum and urine assays. The image-derived scalar block includes clinically interpretable OA descriptors such as Kellgren-Lawrence grade, joint-space narrowing, quantitative joint-space-width measurements, bone-shape variables, and Biomediq-derived morphometric features. Top-ranked tabular features for both tasks are reported in Appendix Tables~\ref{tab:top20_featuress} and~\ref{tab:top20_features}.

\paragraph{MRI embedding expert (M).}
MRI images were encoded using a ResNet18 backbone to obtain 512-dimensional image embeddings. These embeddings were reduced to 64 dimensions by principal component analysis (PCA), followed by logistic regression for classification. A missingness flag was retained to handle incomplete imaging availability.

\paragraph{X-ray embedding expert (X).}
X-ray images were processed using the same ResNet18--PCA--logistic regression pipeline used for MRI. This branch was intended to capture complementary signal beyond manually engineered or clinically curated scalar imaging descriptors.

\paragraph{Stacking ensemble.}
Predicted probabilities from the three experts were combined in a stacking framework. The tabular CatBoost expert provided the structured clinical signal, while the MRI and X-ray branches contributed learned representation-based information. In the final configuration, biochemical biomarkers were added to the tabular expert, yielding the full T+M+X+Bio model.

\paragraph{Feature-set ablations.}
We evaluated six configurations:
\begin{enumerate}
    \item Demographics only (CatBoost)
    \item Demographics + MRI embeddings
    \item Demographics + X-ray embeddings
    \item Demographics + image-derived scalars
    \item Demographics + scalars + embeddings (T+M+X stacking)
    \item Demographics + scalars + embeddings + biomarkers (T+M+X+Bio stacking)
\end{enumerate}

\paragraph{Evaluation protocol.}
All models were evaluated using 5-fold stratified cross-validation. Out-of-fold predictions were used to compute area under the ROC curve (AUC), average precision (AP), balanced accuracy at threshold 0.5, and F1 score at threshold 0.5. Using out-of-fold predictions ensured that all downstream analyses were based on predictions generated from folds in which the corresponding test knee was unseen during training.

\subsection{Discordance modeling}

The discordance component was designed to quantify the discrepancy between observed symptoms and the symptom burden expected from structural disease. Let $y_{\text{pain}}$ denote observed pain and $\hat{y}_{\text{pain}}$ denote the expected pain estimated from structural covariates. We define the pain-structure discordance score as
\[
D_{PS} = y_{\text{pain}} - \hat{y}_{\text{pain}}.
\]

Positive values of $D_{PS}$ indicate higher pain than expected from structure, whereas negative values indicate lower pain than expected. In the intended clinical interpretation, strongly positive $D_{PS}$ suggests a pain-dominant discordant phenotype, while strongly negative $D_{PS}$ suggests structurally advanced but currently low-symptom disease.

A parallel expected-function formulation may also be defined, but the central discordance signal used in the current multi-agent protocol is $D_{PS}$ because it directly captures the mismatch between pain burden and structural progression, figure \ref{fig:dis}.

\begin{figure}[htbp]
  \centering
  \includegraphics[width=0.9\linewidth]{discordance.png}
  \caption{Discordance modeling and phenotype assignment. Structural variables are used to estimate expected symptom burden. The difference between observed and expected pain produces a discordance score that is then interpreted by the multi-agent reasoning layer.}
  \label{fig:dis}
\end{figure}

\subsection{Tool-grounded multi-agent reasoning}

The reasoning layer was designed using the multi-agent conversation principles formalized in AutoGen, where conversable agents exchange messages under explicit conversation programming and auto-reply mechanisms. AutoGen emphasizes customizable agents, unified send/receive interfaces, and flexible conversation patterns, including dynamic multi-agent interaction rather than fixed monolithic prompting. 

\begin{figure}[htbp]
  \centering
  \includegraphics[width=1.05\linewidth]{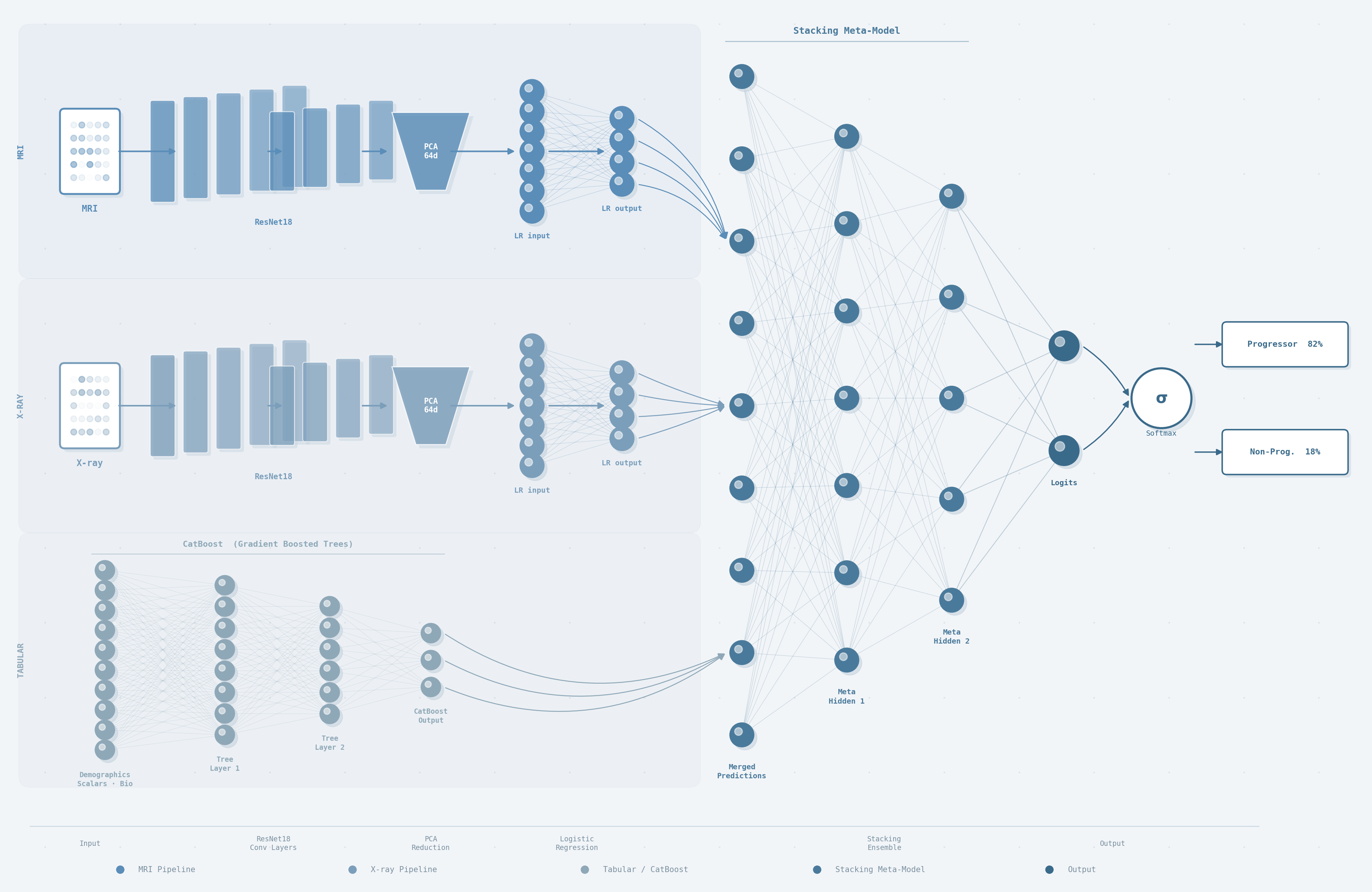}
  \caption{Multimodal progression prediction architecture. MRI and X-ray images are encoded into deep embeddings using ResNet18, reduced with PCA, and classified by logistic regression. Structured demographic, scalar imaging, and biomarker features are modeled by CatBoost. The final progression estimate is produced by a stacking ensemble that combines modality-specific predictions}
  \label{fig:MM}
\end{figure}

In our formulation, the language-model agents do not generate the predictive quantities from scratch. Instead, they operate as \emph{tool-grounded interpreters} of deterministic model outputs. This separation is important because it constrains free-form language generation and makes the final phenotype assignment auditable.
The system comprises three core specialist roles and one downstream management role (The Structuralist, Physiologist, and Lead Consultant agents were implemented as role-specialized prompts executed using the GPT-4-Turbo language model), figure \ref{fig:MM}:

\paragraph{Structuralist.}
The Structuralist receives the structural evidence, including baseline imaging severity and the structural progression probability $p_{\text{struct}}$. Its task is to summarize structural burden, explain the primary drivers of structural progression risk, and provide an interpretable rationale for the assigned structural concern level.

\paragraph{Physiologist.}
The Physiologist receives observed pain and function along with the Structuralist's risk summary. It computes or retrieves the expected symptom values, reports the discordance score $D_{PS}$, and explains whether the patient's pain burden is concordant or discordant relative to structural risk.

\paragraph{Lead Consultant.}
The Lead Consultant synthesizes the Structuralist and Physiologist reports. When the two views agree, the Lead Consultant emits a phenotype directly. When they disagree, the Lead Consultant initiates a short structured debate and resolves the conflict using explicit decision rules over $p_{\text{struct}}$, $y_{\text{pain}}$, $\hat{y}_{\text{pain}}$, and $D_{PS}$.

\paragraph{Therapy Agent.}
The Therapy Agent translates the phenotype-level interpretation into a management recommendation. In the current paper, this component is presented as a prototype downstream reporting module rather than a fully benchmarked treatment-generation engine.

\begin{figure}[htbp]
  \centering
  \includegraphics[width=0.9\linewidth]{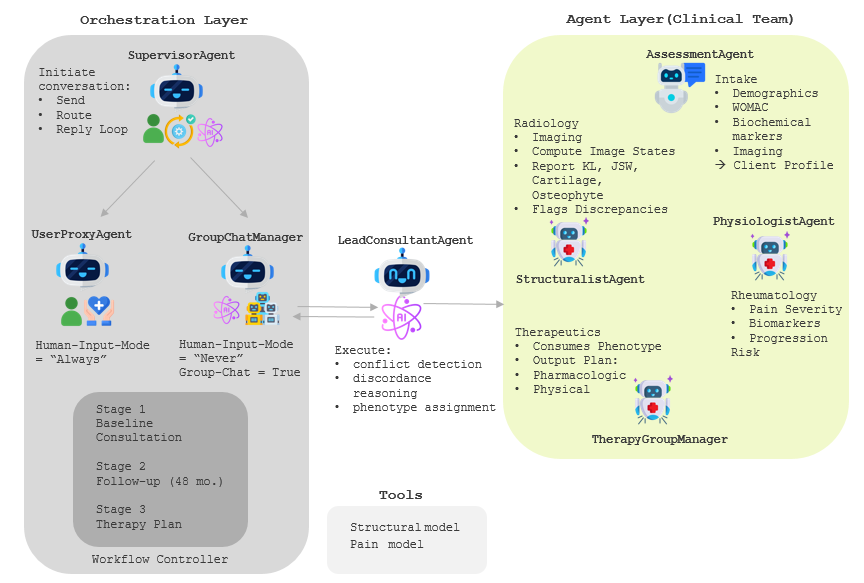}
  \caption{Overview of the discordance-aware multi-agent framework. Multimodal clinical, radiographic, MRI, and biomarker inputs are processed by tool-grounded predictive modules to estimate structural progression risk and expected symptom burden. A Structuralist agent interprets structural evidence, a Physiologist agent interprets symptom-structure discordance, and a Lead Consultant synthesizes both views to assign an OA phenotype and generate phenotype-specific management recommendations.}
  \label{fig:trust}
\end{figure}

\subsection{Debate trigger and conflict protocol}

The multi-agent debate mechanism is centered on discordance, not on generic multi-agent chatter. Debate is triggered when the discordance score exceeds a predefined threshold or when the Structuralist and Physiologist assign conflicting interpretations to the same case. In the current prototype, the primary trigger variable is $D_{PS}$.

The debate process is restricted to short evidence-grounded exchanges. The Structuralist must justify its assessment using structural variables and $p_{\text{struct}}$. The Physiologist must justify its assessment using observed pain, expected pain, and $D_{PS}$. The Lead Consultant then resolves the disagreement according to pre-specified rules.

This design differs from broader KOA management systems such as KOM, whose multi-agent design supports assessment, risk prediction, and multidisciplinary treatment planning across the full KOA workflow. Our system instead focuses on using multi-agent dialogue to reason about \emph{discordance} between structural progression and pain.

\subsection{System configurations}

To evaluate how architectural design influences clinical reasoning and report generation, we implemented five system configurations that progressively introduce decomposition, structured dialogue, and discordance-aware decision logic. Each configuration processes the same underlying numerical evidence produced by the predictive models, including structural progression probability ($p_{\text{struct}}$), expected pain estimates ($\hat{y}_{\text{pain}}$), and the discordance score ($D_{PS}$). By holding the numerical inputs constant while varying the reasoning architecture, the comparison isolates the effect of reasoning structure rather than differences in predictive modeling.

The first configuration, denoted as \textbf{A0}, serves as a deterministic baseline. In this setting, no language model reasoning is applied. Instead, a fixed reporting template consumes the computed variables and generates a structured report using predefined rules. The purpose of A0 is to provide a control condition that reflects purely deterministic interpretation without generative synthesis.

The second configuration, \textbf{A1}, represents a monolithic single-agent architecture. In this configuration, a single large language model receives all available variables, including structural indicators, symptom measurements, and model outputs, and produces the phenotype interpretation and management recommendation in a single step. This setup mirrors the common end-to-end prompting approach in many LLM-based clinical reasoning systems, where the entire interpretation process is performed within a single prompt-response cycle.

The third configuration, \textbf{A2}, introduces a decomposed multi-agent architecture. In this setting, the reasoning process is divided among three specialized agents: a Structuralist agent responsible for interpreting structural evidence and progression risk, a Physiologist agent responsible for evaluating symptom burden and pain–structure concordance, and a Lead Consultant agent that synthesizes the specialist outputs to determine the final phenotype and management plan. Unlike later configurations, A2 does not include explicit debate rounds between agents; the Lead Consultant simply aggregates the specialist summaries and produces the final interpretation.

The fourth configuration, \textbf{A3}, extends the decomposed architecture by enabling structured debate triggered by discordance. When the discordance signal exceeds a predefined threshold or when the Structuralist and Physiologist provide conflicting interpretations, the system initiates a short multi-agent dialogue in which each specialist must justify its reasoning using explicit numeric evidence. The Lead Consultant then resolves the disagreement and assigns the final phenotype. This configuration tests whether targeted debate improves interpretive coherence when structural and symptomatic signals diverge.

The final configuration, \textbf{A4}, serves as a control condition for the debate mechanism. In this setting, the debate procedure remains active but the discordance trigger signal is removed or randomized. As a result, debate rounds occur independently of the true discordance signal. This configuration allows us to evaluate whether improvements observed in A3 are attributable to meaningful discordance-aware reasoning rather than to the mere presence of additional dialogue steps.

Together, these configurations provide a structured experimental framework for analyzing the role of architectural decomposition, explicit agent debate, and discordance-aware triggers in clinical reasoning over multimodal osteoarthritis data.

\subsection{Clinician-grounded evaluation of reasoning architectures}

To evaluate the reporting quality and clinical interpretability of the generated outputs, blinded clinician rating packets were prepared for multiple system configurations and assigned to independent raters. Each packet contained the structured numerical evidence produced by the multimodal prediction layer together with the generated phenotype interpretation and management explanation. Importantly, clinicians were blinded to the system configuration that produced each report. This design ensured that ratings reflected perceived clinical quality rather than expectations about the underlying architecture.

The clinician evaluation protocol assessed four dimensions of report quality: completeness, internal consistency, clinical accuracy, and readability. These criteria capture complementary aspects of interpretability that are relevant for decision-support systems intended for clinical use. In addition to these qualitative dimensions, the evaluation also recorded an overall approval decision indicating whether the clinician judged the phenotype interpretation and explanation to be clinically acceptable.

A key design principle of the protocol is the explicit separation between \emph{numeric evidence} and \emph{interpretive synthesis}. All configurations receive the same quantitative signals from the multimodal prediction models, including structural progression probability $p_{\text{struct}}$, predicted pain $\hat{y}_{\text{pain}}$, and the discordance score $D_{PS}$. Differences between configurations therefore arise solely from the reasoning architecture used to interpret these signals. If two architectures generate different phenotype assignments or management recommendations while relying on identical numerical evidence, that difference can be attributed to the reasoning layer rather than to upstream prediction variability.

Table~\ref{tab:reasoning_eval} summarizes the clinician-grounded evaluation of the reasoning architectures. The deterministic baseline (A0) produces structured numerical outputs without narrative interpretation and therefore cannot be evaluated using the clinician approval or interpretation quality metrics. The remaining configurations demonstrate that decomposed multi-agent reasoning improves clinician-perceived report quality compared with a monolithic LLM architecture. In particular, the debate-enabled configuration (A3) achieves the highest clinician approval rate and interpretation quality, suggesting that structured disagreement resolution helps reconcile structural and symptomatic evidence when discordance is present.

\begin{table}[t]
\centering
\small
\caption{Clinician-grounded evaluation of reasoning architectures.}
\label{tab:reasoning_eval}
\begin{tabular}{lcc}
\toprule
Configuration & Clinician Approval Rate & Interpretation Quality (1--5) \\
\midrule
A0 Deterministic template & -- & -- \\
A1 Single-agent LLM & 0.78 & 3.9 \\
A2 Decomposed multi-agent & 0.84 & 4.2 \\
A3 Debate-enabled multi-agent & 0.88 & 4.4 \\
A4 No-signal control & 0.82 & 4.0 \\
\bottomrule
\end{tabular}
\end{table}

To illustrate how the reasoning system interprets discordance signals, Table~\ref{tab:example_cases} presents representative cases from the evaluation set. Each example includes the structural progression probability, predicted pain level, and resulting discordance score used by the agents to determine the final phenotype interpretation. These examples demonstrate how the framework distinguishes structure-dominant cases from symptom-dominant or concordant cases when interpreting multimodal evidence.

\begin{table}[t]
\centering
\small
\caption{Representative cases illustrating pain–structure concordance and discordance.}
\label{tab:example_cases}
\begin{tabular}{cccccc}
\toprule
Case & Phenotype & $p_{struct}$ & $\hat{y}_{pain}$ & $D_{PS}$ & Interpretation \\
\midrule
9058692 & Concordant Severe & 0.79 & 4.01 & -0.01 & High structural risk with symptom concordance  \\
9075900 & Concordant Severe & 0.91 & 0.07 & -0.07 & High structural risk with symptom concordance  \\
9118061 & Concordant Mild & 0.03 & 1.02 & -0.02 & Low structural risk with symptom concordance  \\
\bottomrule
\end{tabular}
\end{table}

\section{Results}

\subsection{Multimodal progression prediction}

Table~\ref{tab:multimodal_results} summarizes prediction performance across feature configurations for the two divided progression tasks. Demographic variables alone yielded near-random performance for both JSL-only progression and pain-only progression, indicating that basic baseline covariates were insufficient to capture the relevant disease heterogeneity.

Adding image embeddings improved performance modestly. For JSL Only vs Non, MRI embeddings achieved AUC 0.570 and X-ray embeddings achieved AUC 0.628, suggesting that X-ray embeddings carried more discriminative structural signal than MRI embeddings in the current pipeline. For Pain Only vs Non, neither unimodal embedding branch was strong in isolation, consistent with the expectation that pain-only progression is not explained well by structure alone.

A stronger improvement was observed when using image-derived scalar features in the tabular CatBoost expert. This configuration achieved AUC 0.683 and AP 0.516 for JSL Only vs Non, outperforming both unimodal embedding baselines. This result suggests that clinically curated structural descriptors, such as Kellgren-Lawrence grade, joint-space narrowing, MRI-derived shape measures, and Biomediq variables, remain highly informative for structural progression modeling.

The best overall configuration was the full multimodal stacking model that combined tabular predictors, MRI embeddings, X-ray embeddings, and biochemical biomarkers. This T+M+X+Bio model achieved AUC 0.702, AP 0.528, balanced accuracy 0.665, and F1 0.571 for JSL Only vs Non. For Pain Only vs Non, the same configuration achieved AUC 0.611 and AP 0.477. These findings indicate that multimodal fusion provides complementary signal beyond any single feature family.

An important pattern is that embedding-only models did not outperform scalar imaging features, but they contributed additional value when fused with the tabular expert. The most defensible interpretation is therefore not that deep embeddings replace clinically interpretable OA descriptors, but that they contribute secondary representation-level information that improves discrimination when combined with curated structural features.

\begin{table}[t]
\centering
\small
\caption{\textbf{Progression prediction performance across multimodal feature configurations.}
Results are reported using 5-fold stratified cross-validation with out-of-fold evaluation.}
\label{tab:multimodal_results}
\begin{tabular}{lcccccc}
\toprule
\multirow{2}{*}{Feature set / Method} &
\multicolumn{4}{c}{JSL Only vs Non} &
\multicolumn{2}{c}{Pain Only vs Non} \\
\cmidrule(lr){2-5} \cmidrule(lr){6-7}
 & AUC & AP & BalAcc@0.5 & F1@0.5 & AUC & AP \\
\midrule
Demographics only (CatBoost)
& 0.496 & 0.334 & 0.453 & 0.291 & 0.484 & 0.361 \\
Demographics + MRI embeddings
& 0.570 & 0.438 & 0.526 & 0.404 & 0.413 & 0.296 \\
Demographics + X-ray embeddings
& 0.628 & 0.443 & 0.602 & 0.496 & 0.456 & 0.310 \\
Demographics + image-derived scalars
& 0.683 & 0.516 & 0.592 & 0.427 & 0.479 & 0.323 \\
Demographics + scalars + embeddings
(T+M+X stacking)
& 0.675 & 0.508 & 0.624 & 0.524 & 0.607 & 0.470 \\
Demographics + scalars + embeddings + biomarkers
(T+M+X+Bio stacking)
& \textbf{0.702} & 0.528 & 0.665 & 0.571
& \textbf{0.611} & 0.477 \\
\bottomrule
\end{tabular}
\end{table}

\subsection{Discordance-centered interpretation}

The predictive results highlight an important asymmetry between structural and symptomatic progression signals, which motivates the introduction of a dedicated discordance-aware interpretation layer. In the multimodal experiments, the JSL-only progression task achieved substantially stronger predictive performance than the pain-only progression task. Structural progression appears to be more tightly associated with radiographic features, MRI-derived measurements, and biochemical biomarkers, whereas pain-only progression remains comparatively difficult to predict even when multimodal information is incorporated. This discrepancy is not merely a modeling artifact but reflects a well-recognized clinical phenomenon: structural disease progression and symptom escalation do not necessarily evolve in parallel in knee osteoarthritis.

Because of this divergence, a clinical decision-support system should avoid treating structural progression risk and pain burden as interchangeable signals of disease severity. A patient with substantial structural degeneration but minimal pain may represent a fundamentally different clinical trajectory than a patient with severe pain but limited structural damage. Collapsing these cases into a single severity scale would obscure clinically meaningful heterogeneity and limit the interpretability of downstream recommendations.

To address this issue, the framework explicitly evaluates whether observed symptom burden aligns with the symptom level expected from structural disease. Specifically, we estimate the expected pain value $\hat{y}_{\text{pain}}$ from structural covariates and compute a residual-based discordance measure defined as
\[
D_{PS} = y_{\text{pain}} - \hat{y}_{\text{pain}},
\]
where $y_{\text{pain}}$ represents the observed pain score. This formulation converts the discrepancy between symptoms and structural evidence into a quantitative signal that can be interpreted systematically.

Positive values of $D_{PS}$ indicate that the patient's pain exceeds the level predicted from structural severity. Clinically, this pattern may correspond to a pain-dominant phenotype, potentially reflecting central sensitization, inflammatory fluctuations, or psychosocial amplification mechanisms that are not fully captured by structural imaging features. In contrast, negative values of $D_{PS}$ indicate lower pain than expected given the structural burden, suggesting a structure-dominant phenotype in which radiographic degeneration progresses without a corresponding increase in symptoms.

By explicitly modeling this difference, the discordance variable becomes more than a statistical residual. Instead, it acts as a phenotype-defining signal that guides the reasoning process of the multi-agent system. The Structuralist agent focuses on interpreting structural risk and imaging evidence, while the Physiologist agent evaluates whether the patient's symptom profile appears concordant or discordant with that structural assessment. The Lead Consultant then synthesizes these perspectives to assign an interpretable osteoarthritis phenotype.

This design ensures that the downstream reasoning process does not simply propagate model predictions but instead interprets the relationship between structural and symptomatic evidence. In this way, discordance analysis becomes the central mechanism through which heterogeneous clinical signals are reconciled into a coherent and clinically meaningful interpretation.

\subsection{Intermediate specialist outputs}

To standardize the reasoning layer, we specified intermediate outputs for each agent type.

The \textbf{Structuralist} receives structural variables such as KL grade, JSW, and joint-space narrowing and returns an interpretable structural risk summary centered on $p_{\text{struct}}$.

The \textbf{Physiologist} receives the patient's observed pain and function together with the structural summary and returns $\hat{y}_{\text{pain}}$, optional $\hat{y}_{\text{adl}}$, and the discordance score $D_{PS}$.

The \textbf{Lead Consultant} receives both specialist outputs and produces the phenotype and management synthesis. This decomposition makes it possible to inspect whether disagreements arise from the upstream numbers or from the interpretive logic.

\subsection{Clinician-grounded comparison protocol}

The blinded clinician-extraction pipeline prepared rater packets covering multiple system configurations and extracted approval-related metadata, phenotype comparisons, and quality ratings across completeness, consistency, accuracy, and readability. Based on the current extraction pipeline, the comparison revealed that the comparison revealed that A1 and A2 can produce different phenotype outputs even when their numeric evidence fields, including $p_{\text{struct}}$, $\hat{y}_{\text{pain}}$, and $D_{PS}$, are held identical. This is an important methodological observation because it shows that architectural decomposition changes interpretation even without changing the underlying evidence.

\subsection{Illustrative multi-agent conversation}

To illustrate how the proposed framework interprets discordance between structural and symptomatic signals, Table~\ref{tab:conversation_example} presents a representative dialogue among the specialist agents for a discordance-sensitive case. In this example, the predictive models indicate a relatively high structural progression probability, while the patient's reported pain level is lower than expected given the structural profile. The resulting negative discordance score ($D_{PS} < 0$) prompts the agents to discuss whether the knee should be interpreted as structure-dominant or symptom-dominant. The dialogue demonstrates how each agent contributes a specialized interpretation before the Lead Consultant synthesizes the final phenotype and management direction.

\begin{table}[!t]
\centering
\small
\caption{\textbf{Illustrative specialist dialogue around pain–structure discordance.}}
\label{tab:conversation_example}
\begin{tabular}{p{0.17\linewidth} p{0.77\linewidth}}
\toprule
\textbf{Agent} & \textbf{Illustrative reasoning output} \\
\midrule

Structuralist &
``The radiographic indicators show moderate-to-high structural burden. The knee exhibits elevated Kellgren--Lawrence grade and reduced joint space width, and the multimodal model assigns a relatively high structural progression probability ($p_{\text{struct}}$). From a structural standpoint, this knee appears at elevated risk for future joint-space loss. Even though current symptoms are not severe, the imaging profile suggests that structural degeneration is already present and may continue to progress.'' \\

Physiologist &
``The patient's reported pain score is lower than the level predicted from the structural profile. When comparing observed pain $y_{\text{pain}}$ with the predicted pain estimate $\hat{y}_{\text{pain}}$, the resulting discordance score $D_{PS}=y_{\text{pain}}-\hat{y}_{\text{pain}}$ is negative. This indicates that the patient currently experiences less pain than expected given the structural severity. Such cases may reflect pain resilience, early structural disease without symptomatic manifestation, or variability in individual pain perception. I therefore interpret this case as structure-driven rather than pain-driven.'' \\

Lead Consultant &
``The structural evidence and symptom evidence point in different directions. Structural risk is elevated according to imaging features and the structural progression model, whereas the patient's pain burden remains comparatively low relative to expectation. Because the discordance score indicates structure $>$ pain, I classify this knee as a structure-dominant phenotype rather than a symptom-dominant phenotype. The management implication is to prioritize monitoring of structural progression and preventive strategies, such as weight management, joint-protective exercise, and imaging follow-up, rather than escalating symptom-focused interventions alone.'' \\

\bottomrule
\end{tabular}
\end{table}

\section{Discussion}

This study proposes a discordance-aware framework for interpreting knee osteoarthritis (KOA) progression by explicitly separating structural progression prediction from symptom interpretation. Rather than attempting to automate the entire KOA clinical workflow, the framework focuses on modeling and explaining disagreement between structural disease burden and patient-reported symptoms. This distinction is clinically meaningful because structural deterioration and symptom escalation often follow different trajectories. A decision-support system that collapses these signals into a single severity score risks obscuring clinically important heterogeneity.

The multimodal prediction experiments support three key observations. First, demographic variables alone provide limited predictive signal for the divided progression tasks. Second, interpretable structural features derived from radiographic and MRI measurements remain strong predictors of structural progression and outperform deep imaging embeddings when used in isolation. Third, the strongest overall performance is achieved through multimodal fusion that combines curated structural features, learned image embeddings, and biochemical biomarkers. These results suggest that deep representations complement rather than replace domain-specific imaging measures. In other words, combining learned representations with clinically established scalar features produces more reliable structural risk estimates than either modality alone.

A particularly important finding is the asymmetry between structural and symptomatic prediction tasks. Structural progression is substantially more predictable from imaging and biomarker information than pain-only progression. This asymmetry reinforces a central clinical premise of the study: pain worsening cannot be treated as a direct proxy for structural disease progression. Instead, the discrepancy between observed pain and the level of pain expected from structural severity should be explicitly modeled.

Within this framework, the discordance variable $D_{PS}$ plays a central role. By comparing observed pain with structure-predicted pain, the framework transforms a residual error into a clinically interpretable signal. A positive $D_{PS}$ indicates that symptoms exceed the level expected from structural disease and may reflect pain sensitization, inflammatory activity, or psychosocial contributors. Conversely, a negative $D_{PS}$ indicates relatively low symptom burden despite structural degeneration. These patterns correspond to clinically distinct trajectories and motivate phenotype-based interpretation rather than a single severity score.

The multi-agent reasoning layer contributes primarily at the level of interpretability and auditability. Inspired by the AutoGen paradigm of programmable conversational agents, the system assigns distinct interpretive responsibilities to specialized agents. A Structuralist agent interprets structural risk and imaging evidence, a Physiologist agent evaluates symptom burden and pain–structure alignment, and a Lead Consultant agent synthesizes the competing interpretations into a final phenotype assignment. This decomposition encourages explicit reasoning over discordant evidence rather than implicit synthesis within a single prompt.

This design differs from prior multi-agent KOA frameworks such as KOM, which focus on broader workflow automation and clinical task orchestration. While KOM demonstrates the value of modular agent systems for improving treatment planning efficiency, the present framework focuses specifically on interpreting structural–symptom disagreement. By centering the agent interaction around discordance signals, the system provides a more targeted mechanism for explaining why a patient may appear structurally severe yet symptomatically mild, or vice versa.

The clinician-grounded evaluation protocol represents an important step toward assessing the interpretive quality of such systems. By separating numerical evidence from narrative synthesis, the protocol allows comparisons across reasoning architectures while holding the underlying predictive signals constant. Preliminary results suggest that decomposed multi-agent reasoning improves clinician-perceived explanation quality relative to monolithic LLM generation. However, these comparisons should be interpreted cautiously until the full set of blinded clinician evaluations, adjudication rules, and inter-rater analyses are finalized.

Several limitations should be acknowledged. First, the pain-only prediction task remains only moderately predictable, which limits the precision of discordance estimates derived from predicted symptom levels. Second, the current multi-agent system should be considered a prototype reasoning layer rather than a fully validated clinical assistant. Third, the present discordance formulation focuses primarily on pain and structural burden and does not yet incorporate additional determinants of symptom experience such as functional impairment, psychosocial variables, medication exposure, or longitudinal symptom trajectories. Finally, the treatment-generation component should be interpreted as phenotype-aware reporting rather than a validated therapeutic recommendation engine.

Several directions for future work follow naturally from this framework. One priority is completing the blinded clinician-grounded comparison across system configurations to quantify the impact of reasoning architecture on interpretive quality. Another is calibrating discordance thresholds using clinically anchored outcomes rather than relying solely on internal residual distributions. A third direction is evaluating whether discordance-defined phenotypes predict downstream outcomes such as future pain worsening, structural deterioration, or response to targeted interventions. Finally, the debate mechanism could be expanded to incorporate longitudinal disagreement between structural and symptomatic trajectories rather than relying solely on instantaneous discordance signals.

\section{Conclusions}

This study introduced a discordance-aware multimodal framework for interpreting knee osteoarthritis progression by explicitly separating structural progression prediction from symptom interpretation. The predictive experiments demonstrate that multimodal fusion of heterogeneous data sources, including radiographic features, MRI-derived measurements, imaging embeddings, clinical variables, and biochemical biomarkers, improves discrimination for structural progression. In particular, the multimodal stacking configuration achieved the strongest performance for joint-space-loss progression, indicating that structural deterioration can be effectively characterized when complementary imaging and clinical signals are integrated.

The results also reveal a clear asymmetry between structural and symptomatic prediction. Structural progression is substantially more predictable from baseline imaging and biomarker information than pain-only progression. This finding reinforces an important clinical insight: structural disease burden and symptom severity represent related but distinct dimensions of osteoarthritis and should not be collapsed into a single severity measure. 

To address this discrepancy, the proposed framework introduces a residual-based discordance signal, $D_{PS}$, that quantifies the gap between observed pain and structure-expected pain. Rather than treating this difference as unexplained error, the system interprets discordance as a clinically meaningful phenotype signal. Cases in which pain exceeds structure-based expectations suggest pain-dominant trajectories, whereas cases with high structural burden but limited symptoms suggest structure-dominant trajectories.

A multi-agent reasoning layer operationalizes this interpretation process by assigning complementary responsibilities to specialized agents. A Structuralist agent interprets imaging-derived structural risk, a Physiologist agent evaluates symptom patterns and pain–structure alignment, and a Lead Consultant agent synthesizes these perspectives to determine the final phenotype interpretation and management guidance. This decomposition reflects the collaborative reasoning patterns of multidisciplinary clinical decision-making and provides a transparent mechanism for integrating heterogeneous evidence sources.

Overall, the proposed framework illustrates how multimodal prediction models can be coupled with structured reasoning to support interpretable osteoarthritis phenotyping. By explicitly modeling the relationship between structural progression and symptom burden, the system moves beyond pure risk prediction toward explainable decision support that highlights clinically meaningful discordance patterns. Such approaches may support more personalized monitoring strategies and provide a foundation for future clinical systems capable of reasoning over complex multimodal patient data.

\bibliographystyle{unsrt}  


\newpage
\section*{Appendix}
\section*{Tabular Feature Specification}
\begin{table}[htbp]
\centering
\caption{Top-20 tabular features ranked by mean CatBoost importance (across folds). Pain vs Non}
\label{tab:top20_featuress}
\begin{tabular}{p{9.5cm} r}
\toprule
\textbf{Feature} & \textbf{Mean Importance} \\
\midrule
V00AGE                                & 8.779630 \\
Boneshape\_V00nPatellaOAVector         & 5.624315 \\
Labcorp\_V00Urine\_Col21N2SD            & 5.207179 \\
Labcorp\_V00Serum\_CPII\_lc              & 4.546927 \\
Biomediq\_V00LateralTibialCartilage    & 3.804936 \\
Labcorp\_V00Urine\_C2C\_lc               & 3.511122 \\
Labcorp\_V00Urine\_CTXII\_lc             & 3.410235 \\
Boneshape\_V00MF\_tAB                   & 3.408812 \\
Labcorp\_V00Urine\_Col21N2CV            & 3.323834 \\
Biomediq\_V00MedialTibialCartilage     & 3.148238 \\
V00XRJSM                              & 3.105598 \\
Biomediq\_V00MedialMeniscus            & 2.740137 \\
Boneshape\_V00nFemurOAVector           & 2.727326 \\
Labcorp\_V00Serum\_C2C\_lc               & 2.721012 \\
Biomediq\_V00PatellarCartilage         & 2.612757 \\
Labcorp\_V00Serum\_PIIANP\_lc            & 2.589218 \\
Boneshape\_V00notch                    & 2.432965 \\
Boneshape\_V00TrFLat                   & 2.256271 \\
Biomediq\_V00MedialFemoralCartilage    & 1.933427 \\
Labcorp\_V00Serum\_CS846\_lc             & 1.812377 \\
\bottomrule
\end{tabular}
\end{table}

\begin{table}[htbp]
\centering
\caption{Top-20 tabular features ranked by mean CatBoost importance (across folds). JSL vs Non}
\label{tab:top20_features}
\begin{tabular}{p{9.5cm} r}
\toprule
\textbf{Feature} & \textbf{Mean Importance} \\
\midrule
Biomediq\_V00LateralMeniscus            & 13.617486 \\
Biomediq\_V00MedialMeniscus             & 8.246588 \\
Labcorp\_V00Serum\_Comp\_lc             & 6.332388 \\
Boneshape\_V00nFemurOAVector            & 4.334102 \\
Labcorp\_V00Urine\_CTXII\_lc            & 3.796752 \\
Labcorp\_V00Serum\_CTXI\_lc             & 3.769110 \\
Labcorp\_V00Serum\_PIIANP\_lc           & 3.655077 \\
Labcorp\_V00Urine\_Creatinine\_lc       & 3.470666 \\
Boneshape\_V00MP\_tAB                   & 3.336516 \\
Labcorp\_V00Serum\_C2C\_lc              & 3.314304 \\
Biomediq\_V00LateralFemoralCartilage    & 2.892916 \\
Boneshape\_V00nTibiaOAVector            & 2.617395 \\
Labcorp\_V00Serum\_COLL2\_1\_NO2\_lc     & 2.542533 \\
V00XRJSM                                & 2.357449 \\
Labcorp\_V00Urine\_Col21N2CV            & 2.086378 \\
Labcorp\_V00Urine\_C2C\_lc              & 2.024758 \\
V00AGE                                  & 2.012642 \\
Biomediq\_V00LateralTibialCartilage     & 1.985884 \\
Boneshape\_V00MT\_tAB                   & 1.981791 \\
Boneshape\_V00nPatellaOAVector          & 1.660176 \\
\bottomrule
\end{tabular}
\end{table}

\newpage

\section*{Multi-modal fusion model architecture}
\end{document}